\documentclass[10pt,twocolumn,letterpaper]{article}

\usepackage{wacv} 
\usepackage[accsupp]{axessibility} % Improves PDF readability for those with disabilities.
\usepackage{graphicx}
\usepackage{amsmath}
\usepackage{amssymb}
\usepackage{booktabs}

\usepackage{tikz}
\usepackage{pifont}
\usepackage{multirow}

\usepackage[pagebackref,breaklinks,colorlinks]{hyperref}

\usepackage[capitalize]{cleveref}
\crefname{section}{Sec.}{Secs.}
\Crefname{section}{Section}{Sections}
\Crefname{table}{Table}{Tables}
\crefname{table}{Tab.}{Tabs.}

\def\wacvPaperID{2018} % *** Enter the WACV Paper ID here
\def\confName{WACV}
\def\confYear{2025}

\author{Luca Scofano* \quad Alessio Sampieri*\quad Edoardo De Matteis*\\ 
\quad Indro Spinelli \quad Fabio Galasso \\\\
{\tt\small \{scofano, sampieri\}@diag.uniroma1.it \quad \{dematteis,  spinelli, galsso\}@di.uniroma1.it} \\\\ Sapienza University of Rome, Italy} 

\begin{document}

\title{Social EgoMesh Estimation}

\maketitle
\def\thefootnote{*}\footnotetext{Authors contributed equally.}

\begin{abstract}
Accurately estimating the 3D pose of the camera wearer in egocentric video sequences is crucial to modeling human behavior in virtual and augmented reality applications. 
The task presents unique challenges due to the limited visibility of the user's body caused by the front-facing camera mounted on their head. 
Recent research has explored the utilization of the scene and ego-motion, but it has overlooked humans' interactive nature.
We propose a novel framework for \textbf{S}ocial \textbf{E}gocentric \textbf{E}stimation of body \textbf{ME}shes (SEE-ME).
Our approach is the first to estimate the wearer's mesh using only a latent probabilistic diffusion model, which we condition on the scene and, for the first time, on the social wearer-interactee interactions.
Our in-depth study sheds light on when social interaction matters most for ego-mesh estimation; it quantifies the impact of interpersonal distance and gaze direction.
Overall, SEE-ME surpasses the current best technique, reducing the pose estimation error (MPJPE) by 53\%.
The code is available at \href{https://github.com/L-Scofano/SEEME/tree/main}{SEEME}.
\end{abstract}    
\section{Introduction}
\label{sec:intro}

\begin{figure*}[h!t]
    \centering
    \subfloat[Wearer's first person view, looking at the interactee.]{
        \includegraphics[width=.45\textwidth, bb=0 0 1900 1000]{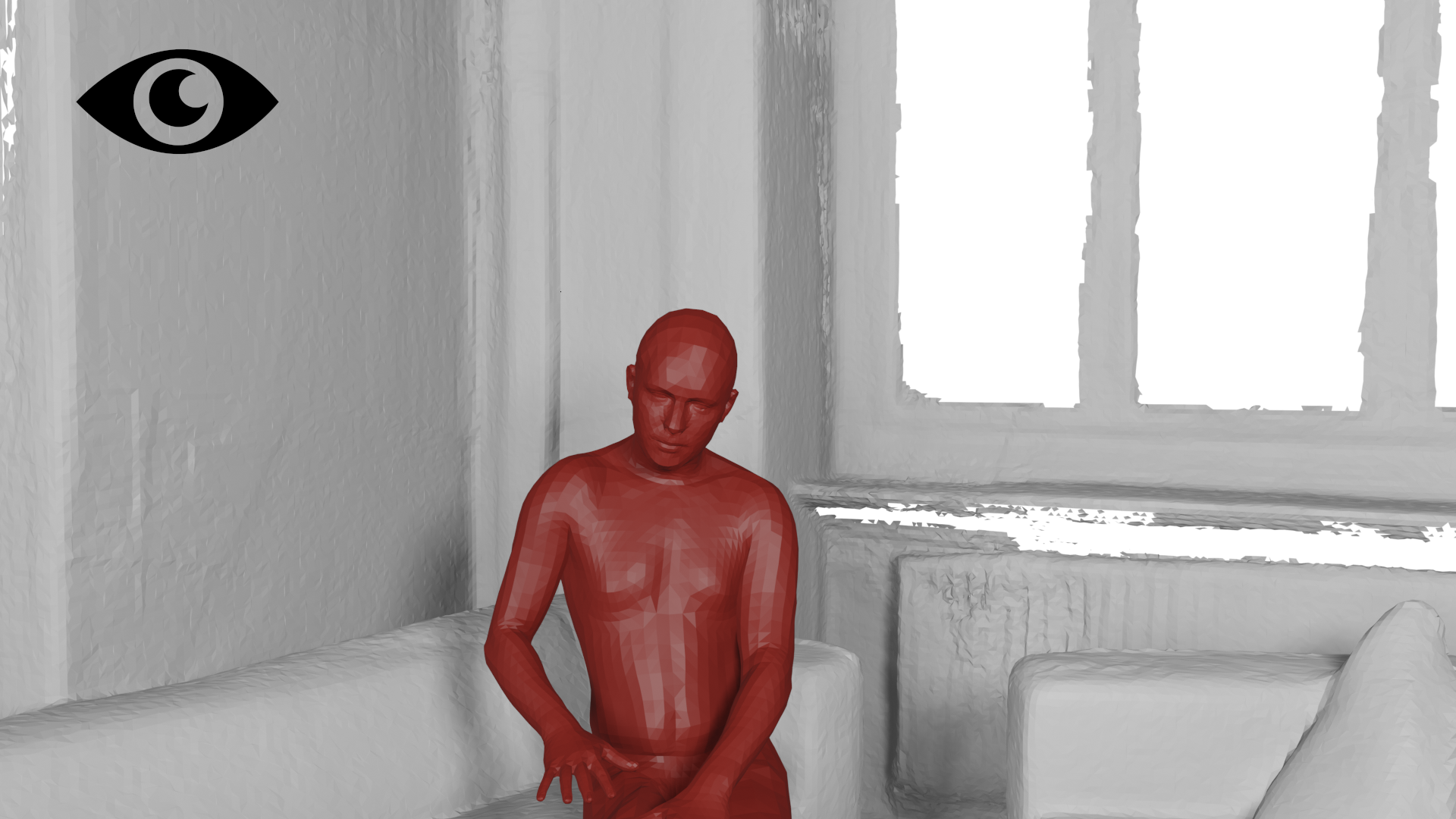}
        \label{fig:teaser:first}
    }
    \hfill
    \subfloat[Third-person view of the wearer and the interactee in a scene.]{
        \includegraphics[width=.45\textwidth, bb=0 0 1900 1000]{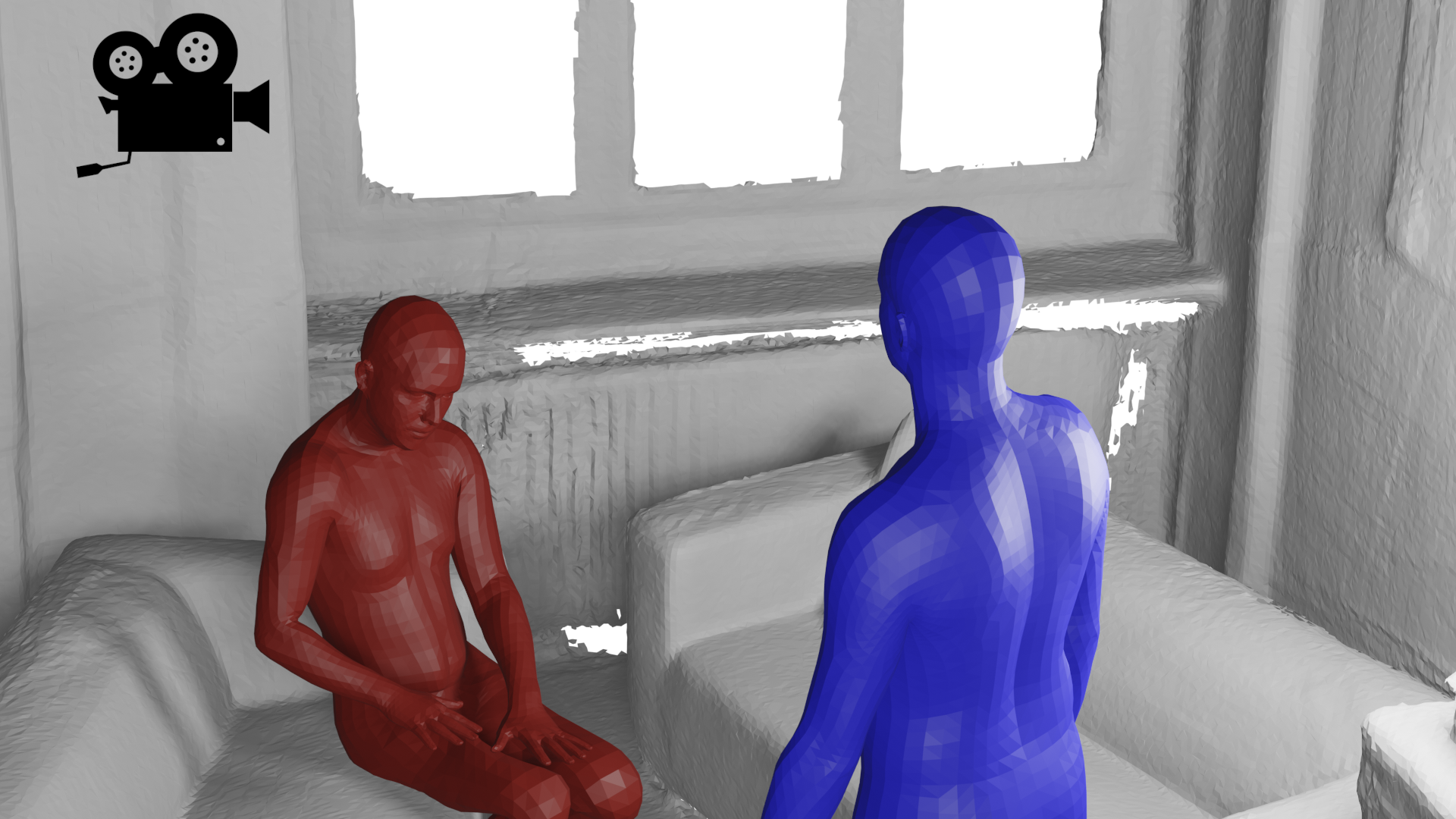}
        \label{fig:teaser:third}
    }
    \caption{
    \textit{(Left)} Frame from the input egocentric video stream. We experience the immersive subjective perspective of the front-facing camera wearer, but the wearer is behind the wearable and, therefore, invisible. Still, we recognize the points of interest of the wearer, parts of the scene where the action happens, and, most importantly, the interactee engaged in communication with the wearer. \textit{(Right)} Third-view reconstruction of the \textit{ego} mesh of the camera wearer by our proposed \textbf{SEE-ME}. Our vantage point reveals the surrounding environment, featuring a sofa and a person from an overhead perspective, leading us to infer the wearer's likely standing position. Specifically, vicinity and gaze interactions are important cues for our reconstruction as we experimentally quantify.
    }
    \label{fig:teaser}

\end{figure*}

% \begin{figure}[!tp] 
%     \centering
%     \subfloat[Length-aware latent space.]{%
%         \includegraphics[width=0.38\textwidth]{imgs/eccv_teaser_pov_glyph.png}%
%         \label{fig:lat_spaces}%
%         }%
%     \hfill%
%     \subfloat[Time/performances trade-off.]{%
%         \includegraphics[width=0.52\textwidth]{imgs/eccv_teaser_glyph.png}%
%         \label{fig:bin}%
%         }%
%     \caption{Length aware in-depth analysis. Zoom in for details.}
%     \vspace{-.3cm}
% \end{figure}

Estimating the 3D motion of a person from an egocentric video sequence is a critical task with diverse applications in virtual reality (VR) and augmented reality (AR). 
This research is motivated by the necessity to realistically depict the entire body to enable total immersion in these environments~\cite{Chessa2019GraspingOI}. 
The user typically wears a camera in such videos, capturing their surroundings from a first-person perspective. 
Large field of view (FOV) cameras can capture some body parts (hands, feet), which standard FOV cameras cannot, as illustrated from the first-person perspective in Figure~\ref{fig:teaser}.

The task, dubbed egopose estimation, is influenced by the type of camera used to record the sequences. Top-down head-mounted cameras have the best view of the wearer \cite{tome19xr} but are intrusive due to the required displacement from the face. Front-facing cameras solve the ergonomic issue, but the downside is that the wearer is almost always invisible except when using very large FOV. 

In \cite{Jiang2021EgocentricPE}, the authors introduce a two-stage framework for fish-eye head-mounted cameras, where a traditional SLAM algorithm \cite{shinya19openslam} provides the camera rotation and translation. 
Simultaneously, a neural network exploits visible body parts to predict the egopose.
Recently, EgoEgo~\cite{Li2022EgoBodyPE} proposed an updated pipeline, relaxing the assumption on the large FOV. 
However, they still rely on (deep learning-based) SLAM \cite{teed2021droid} to estimate the head's rotation and translation, fed to a generative model \cite{Ho2020DenoisingDP} to synthesize the pose for the rest of the body. 
These approaches exploit the video to extract the head pose, ignoring the surrounding environment and the other actors in the scene that can provide essential cues for the egopose estimation. 

For this reason, we propose a new social egocentric mesh estimation task and contribute with a novel probabilistic framework for \textbf{S}ocial \textbf{E}gocentric \textbf{E}stimation of body \textbf{ME}shes (\textbf{SEE-ME}). 
Humans have a social nature and are often involved in social activities in real and virtual worlds, accomplishing tasks that usually require cooperation and coordination.
You2Me \cite{Ng2019You2MeIB} explicitly captures the interplay between two persons from a chest-mounted camera in a controlled environment where two people interact by conversing, playing hand games, or doing sports. 
However, their approach does not extract high-level scene information and predicts deterministic skeleton-like representation that may not be suitable for AR/VR applications.
In contrast with our work, \cite{Ng2019You2MeIB} employs a person-centric coordinate system and does not estimate global orientation or translation, while our approach enables end-to-end learning of both of them. 
\cite{Ng2019You2MeIB} also relies on OpenPose~\cite{cao2017realtime} to extract the interactee, which faces challenges in accurately estimating occluded keypoints~\cite{Liu2023EgoHMREH}. 
We exploit EgoHMR~\cite{Zhang2023ProbabilisticHM} to estimate the interactee's mesh, demonstrating robustness against occlusions, using scenes as input.
%It is important to highlight how \cite{Zhang2023ProbabilisticHM} utilizes scenes as input, serving a different purpose: to recover the missing body joints of a visible person. 
In contrast, our approach leverages the scene and interactions to understand the invisible camera wearer's pose, with no input body cues required.

The motivation behind our research stems from the wearer's perspective, which highlights both their surroundings and an overlooked aspect: the presence of another actor within the scene (see Figure \ref{fig:teaser}). 
Our unified approach links the generation of the egomesh to both the 3D scene depiction and the wearer's social interaction with the interactee.
We do not rely on SLAM for the head's orientation or translation but predict them with our model. 
Our approach pioneers conditional Latent Diffusion in egocentric human mesh estimation, which provides a notable speedup in generation times. 
We build the latent space encoding human poses using a Variational Autoencoder (VAE)~\cite{Kingma2013AutoEncodingVB}.
Subsequently, we perform conditional diffusion on this latent space. 
The conditioning strategies aim to guide the process by modeling the 3D point cloud of the scene in which the wearer moves and the estimated mesh of the interactee recovered from the egocentric video feed.
Our model depends on the 3D scene and other actors' poses. 
Therefore, provided the wearer's pose, we can predict the interactee's one even if it is not paired with an egocentric video stream.

To assess the performance of our approach, we evaluate it on EgoBody~\cite{Zhang2021EgoBodyHB}: the solely available dataset that features multiple individuals, an egocentric perspective, and an environment.
We reach state-of-the-art results, establishing the efficacy of environmental and social components.
In our research, we thoroughly examine the influence of social interactions on estimating the wearer's egomesh. 
We establish when this conditioning strategy has the highest impact. 
We use proxies for social interaction that can be extracted from our setup. 
We demonstrate how proximity and eye contact between the wearer and the interactee bring the most from our conditioning. 
Moreover, we study the effect of future knowledge about the interactee's motion, as we humans use the experience to predict and react to the future movements of the people around us.
Our framework, exposed to this information, provides a significant performance boost.
%\LS{Add that we also test it on GIMO and report improvement on performance.}
To corroborate the strength of our approach, we evaluate our model on the dataset GIMO~\cite{Zheng2022GIMOGH}, which is egocentric and includes the environment but is not multi-person.

To summarize our contributions:
\begin{itemize}
\item We propose the task of social egomesh estimation. Emphasizing the role of other actors in 3D scenes to account for the lack of information about the camera wearer.
% \item We introduce SEE-ME, a monolithic framework based on latent diffusion capable of predicting both the wearer and interactee poses with SOTA performances
\item We introduce SEE-ME, a unified framework based on latent diffusion that achieves state-of-the-art performance in predicting the wearer’s pose and is also capable of predicting the interactee’s pose.
\item We perform an in-depth ablation study to highlight the scenarios in which modeling social interactions brings the most benefits.
\end{itemize}

\section{Related Work}
\label{sec:related}

\paragraph{Pose Estimation from third-person cameras.}
3D pose estimation from images and videos in a third-person perspective has seen significant research efforts in recent years. It can be broadly categorized into two main approaches. The first approach aims to predict the positions of joints from images and videos directly~\cite{Mehta2017VNectR3, Pavlakos2016CoarsetoFineVP, Tung2017SelfsupervisedLO, Zhou2015SparsenessMD, Kocabas2019VIBEVI, Luo20203DHM}. The second approach employs a parametric human body model~\cite{Loper2015SMPLAS} to estimate the parameters of the body model based on images or videos~\cite{ye2023slahmr, Choi2020BeyondSF, Kanazawa2017EndtoEndRO, Kocabas2019VIBEVI, Kolotouros2019LearningTR, Luo20203DHM}. 
Recent approaches have advanced the realism of human motion modeling by incorporating human dynamics. This is achieved through the utilization of learned priors~\cite{Rempe2021HuMoR3H} or physics-based priors~\cite{Grtner2022TrajectoryOF, Peng2018SFVRL, Rempe2020ContactAH, Shimada2021NeuralM3, Xie2021PhysicsbasedHM, Yuan2021SimPoESC}. These priors are inherently formulated within the human coordinate frame. 

However, a notable challenge arises when dealing with egocentric videos, which we consider in this study, where the whole body is often not visible because body joints are mostly hidden from view. By considering the physical and social aspects, we aim to provide we aim to overcome this lack of information.

%\FG{the rest of the paragraph is stating our contribution, not relating to other work, maybe remove it}Our study considers the physical environment and emphasizes the social context, particularly the interactions and dynamics involving the other person in the scene. This holistic approach allows us to comprehensively understand the situation and how individuals relate to their surroundings and each other. By considering the physical and social aspects, we aim to provide a more nuanced and insightful analysis of the context under study.

\paragraph{Motion Estimation from Egocentric Video.}
There is a growing emphasis on pose estimation from egocentric videos, where various hardware setups are utilized, including fisheye cameras~\cite{Hori2022Silhouettebased3H, Tom2020SelfPose3E, Akada2022UnrealEgoAN, Wang2022SceneAwareE3, Liu2023EgoHMREH}, outward-facing cameras~\cite{Shiratori2011MotionCF, Yi2023EgoLocateRM, Jiang2016SeeingIP, Ng2019You2MeIB, Zhang2023ProbabilisticHM, Zhang2021EgoBodyHB, Zheng2022GIMOGH}, and additional inputs such as controllers and synchronized headsets \cite{Zheng2023RealisticFT, Jiang2022AvatarPoserAF}, all aimed at estimating a person's pose. 
None of them consider social interactions. However, there has been a growing interest in incorporating interactions between people in a given scene, as evidenced by numerous studies~\cite{Khirodkar2023EgoHumansAE, Zhang2021EgoBodyHB, Zhang2023ProbabilisticHM, Ng2019You2MeIB}. 
Notably,~\cite{Zhang2021EgoBodyHB, Zhang2023ProbabilisticHM} both focus on estimating the interactee without relying on any additional cues from the camera wearer. 
In contrast, You2Me~\cite{Ng2019You2MeIB} predicts full-body wearer motions by observing the interaction poses of a second person in the camera view. 
While this approach is effective for keypoint estimation, it lacks critical information necessary for our task, such as body meshes, 3D scene details, and global rotation and orientation. 
Conversely, our approach involves probabilistic human mesh estimation, incorporating scene information.

In~\cite{Li2022EgoBodyPE}, a distinctive method is proposed for multi-hypothesis human mesh estimation. 
This approach decouples the problem by estimating head movement, employing it as a conditioning variable for a DDPM~\cite{Ho2020DenoisingDP} model to generate plausible poses.
In contrast, our approach eliminates the reliance on a three-block process involving SLAM~\cite{teed2021droid}, Gravitynet, and Headnet~\cite{Li2022EgoBodyPE}, tackling the problem as a unified mesh estimation task. 
%\FG{if there is any "unfairness" in the comparison with ours, this is a place to state it. Does SLAM imply the consideration of more frames, beyond those which SEE-ME ingests? else ok as is} 
We employ a Latent Diffusion Model to generate the entire body without preprocessing steps such as optical flow estimation and camera localization.
Furthermore, our approach incorporates the social component as a conditioning factor.

\paragraph{Probabilistic models for human pose estimation.}
Due to limited image or video observations and inherent depth ambiguity, estimating a 3D human pose from a single image can give rise to numerous potential solutions, mainly when body truncation is a factor. 
Recent research endeavors have approached this challenge by framing it as a generative process or predicting multiple hypothetical poses. 
In various studies, a discrete set of hypotheses has been generated to address this issue, as seen in~\cite{Oikarinen2020GraphMDNLG, Li2022EgoBodyPE, Zhang2023ProbabilisticHM, Biggs20203DMF, Jahangiri2017GeneratingMD, Li2019GeneratingMH, Kolotouros2021ProbabilisticMF, Wehrbein2021ProbabilisticM3, Liu2023EgoHMREH}.
We extend the advancements made in recent motion diffusion models~\cite{Tevet2022HumanMD, Chen2022ExecutingYC, Zhang2022MotionDiffuseTH, Dabral2022MoFusionAF, Chang2022UnifyingHM}. 
Our approach further integrates conditioning on bystanders or individuals within the scene. 
It is crucial to highlight that only~\cite{Zhang2023ProbabilisticHM} considers the 3D scene constraint. 
Our methodology uses the latent diffusion process to capture the inherent ambiguity in pose estimation, incorporating flexible scene and social conditioning techniques.

\section{Methodology}
\label{sec:methodology}

\begin{figure*}[!htp]
  \centering
  \includegraphics[width=\textwidth, bb=0 0 800 300]{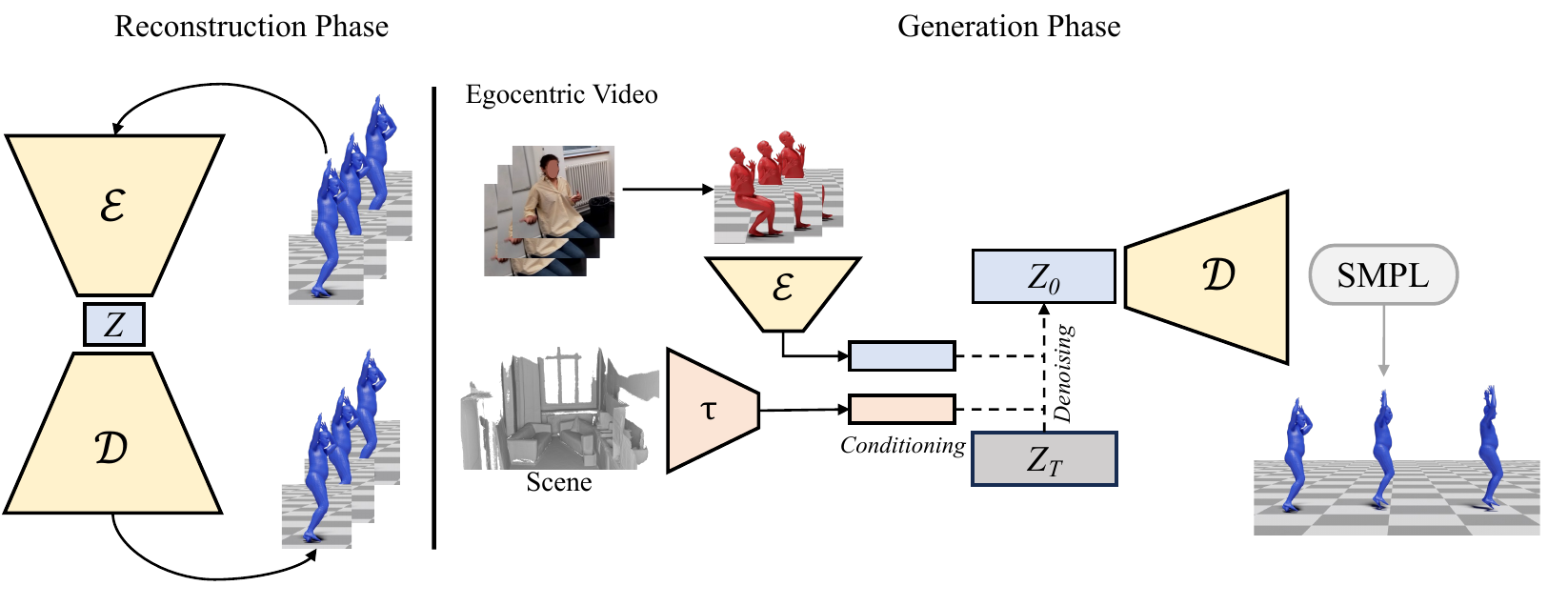}
  \caption{SEE-ME framework. On the left, we present VAE's training to learn a meaningful latent space by solving reconstruction tasks. On the right, we extract and process our conditioning strategies. Corresponding to the 3D point cloud representation of the scene and the interactee's pose extracted from the video sequence. After the conditional denoising process, we can output a SMPL representation of the wearer's pose. }
  \label{fig:model_sketch}
\end{figure*}

Modeling social interaction requires solving a series of challenging tasks. 
    Note, as done in \cite{Li2022EgoBodyPE, Zhang2021EgoBodyHB}, we pose the origin of our coordinate system as the camera of the wearer.
First, we extract a 3D representation of the interactee's human body from a video stream. 
Subsequently, we need to position this representation into a depiction of the 3D environment where the interaction occurs. 
Hence, this is the information that SEE-ME exploits to recover the camera wearer's mesh using only the egocentric video stream, where the wearer itself is not visible. 
We showcase our pipeline in Figure~\ref{fig:model_sketch}. 
Before using social interaction and scene description to drive the generation process, we train part of our model to encode human motion in a latent space (Sec.~\ref{sub3:vae}).
This dramatically reduces the dimensionality required to represent a person.
In this space, we learn to synthesize human motion from pure noise using latent diffusion processes (Sec.~\ref{sub3:posegen}). 

\paragraph{Problem formalization.} Our objective is to generate a plausible representation of the camera wearer $\mathbf{P^w}$, conditioned on the other person in the scene $\mathbf{P^i}$ and the 3D scene $\mathbf{S}$.
We define the sequence of poses of the camera wearer as $\mathbf{P^w}= \{\mathbf{p^w}_k\}_{k=1}^{F}\in \mathbb{R}^{F\times V}$, where $F$ and $V$ are the number of frames and the parameters of the pose vector. Similarly, $\mathbf{P^i}= \{\mathbf{p^i}_k\}_{k=1}^{F}\in \mathbb{R}^{F\times V}$. Following literature~\cite{Chen2022ExecutingYC, Zhang2023GeneratingHM, Lou2023DiverseMotionTD, Zhang2022MotionDiffuseTH}, our poses vectors $\mathbf{P^w}$ and $\mathbf{P^i}$ contain pose, translation and rotation and represents the SMPL~\cite{Loper2015SMPLAS} body representation. Finally, $\mathbf{S}$ is defined as a 3D scene point cloud, denoted as $\mathbf{S} \in \mathbb{R}^{N \times 3}$, where $N$ is the number of points.

\subsection{Latent Human Representation}
\label{sub3:vae}

VAEs consist of an encoder-decoder generative architecture trained to minimize the reconstruction error. 
We employ it to reduce the pose vector $V$ dimensionality, projecting onto a manifold of feasible poses.
In this framework, the encoder network $\mathcal{E}$, parameterized by $\phi$, generates lower-dimensional embeddings $\mathbf{z} \in \mathbb{R}^{n \times D}$ as outputs based on the input poses $\mathbf{P} = \{{\mathbf{p}_k}\}_{k=1}^{F}$.
Following VAE literature \cite{Kingma2013AutoEncodingVB}, $q_\phi(\mathbf{z} \vert \mathbf{p})$ approximates the true posterior distribution of the latent space with a multivariate Gaussian with a diagonal covariance structure:
\begin{equation}
q_\phi(\mathbf{z} \vert \mathbf{p}) = \mathcal{N}(\mathbf{z} \vert \mu_\phi(\mathbf{p}), \sigma^2_\phi(\mathbf{p})\mathbf{I}),
\end{equation}
where $\mu_\phi(\cdot)$ and $\sigma^2_\phi(\cdot)$ are the encoder's outputs. We sample from the approximated posterior $\mathbf{z}_i \sim q_\phi(\mathbf{z}\vert \mathbf{p}_i)$ using:
\begin{equation}
\label{eq:sampling}
\mathbf{z}_i = \boldsymbol{\mu}_i + \boldsymbol{\sigma}_{i}^2 \odot \boldsymbol{\rho},
\end{equation}
where $\mathbf{z}_i$ is a one-dimensional vector of size $D$, and $\boldsymbol{\rho}$ is sampled from a standard multivariate Gaussian distribution. The decoder network $\mathcal{D}$ parametrized by $\theta$ maps the sampled values back to body poses  $p_\theta(\mathbf{p} | \mathbf{z})$ mapped into body meshes with the differentiable SMPL model. The network parameters are obtained by optimizing the Evidence Lower Bound (ELBO) objective, as described in~\cite{Kingma2013AutoEncodingVB}.
The motion decoder $\mathcal{D}$ relies on a transformer decoder architecture ~\cite{Chen2022ExecutingYC, Petrovich2022TEMOSGD} with a cross-attention mechanism, taking $f^*$ zero motion tokens as queries, where $f^*$ is the target sequence length, and a latent $\mathbf{z} \in \mathbb{R}^{1 \times D}$ as memory, ultimately generating a human motion sequence $\mathbf{\hat{p}}_{1:{f^*}}$.  

\subsection{Ego-Mesh Estimation via Latent DDPMs}
\label{sub3:posegen}
We utilize the latent-DDPM framework introduced in~\cite{Rombach2021HighResolutionIS} and adapted for human motion synthesis in ~\cite{Chen2022ExecutingYC} where the diffusion process occurs on a condensed, low-dimensional motion latent space.
Latent DDPM have a different approximated posterior $g(\mathbf{z}_t \vert \mathbf{z}_{t-1})$, denoted \textit{diffusion process}, which gradually gradually converts latent representations $\mathbf{z}_{0}=\mathbf{z}$ into random noise $\mathbf{z}_{T}$ in $T$ timesteps: 
\begin{equation}
    g(\mathbf{z}_t \vert \mathbf{z}_{t-1}) = \mathcal{N}(\mathbf{z}_t; \sqrt{\bar{\alpha}_t}\mathbf{z}_{t-1}, (1-\bar{\alpha}_t)\mathbf{I})\,,
    \label{eq:diffusion}
\end{equation}
where $\bar{\alpha}_t$ is a scaling factor specific to timestep $t$.
Then, the reverse process dubbed \textit{denoising} gradually refines the noised vector to a suitable latent representation $\mathbf{z}_{0}$. 
Following \cite{Chen2022ExecutingYC, Rombach2021HighResolutionIS, Ho2020DenoisingDP, Dhariwal2021DiffusionMB}, we use the notation $\{\mathbf{z}_t\}^T_{t=0}$ to denote the sequence of noised latent vectors, with $\mathbf{z}_{t-1} = \epsilon_\psi(\mathbf{z}_t, t)$ representing the denoising operation at time step $t$.
Here, $\epsilon_\psi$ refers to a denoising autoencoder trained to predict the denoised version of its input.
\begin{equation}
Loss := \mathbb{E}_{\epsilon \sim \mathcal{N}(0,\mathbf{I}),t} [\| \epsilon - \epsilon(\mathbf{z}_t, t) \|_2^2] \,.
\end{equation}
During the denoiser's training, the encoder and decoder remain frozen. 
In the subsequent diffusion reverse stage, $\epsilon_\psi(\mathbf{z}_t, t)$ predicts $\hat{\mathbf{z}}_0$ through a series of $T$ iterative denoising steps. 
Following this, the decoder $\mathcal{D}$ translates $\hat{\mathbf{z}}_0$ into poses $\mathbf{P^i}$ and meshes in a single forward pass.

\subsection{Social Conditioning}
\label{sub3:cond}

We introduce a novel conditioning strategy focused on embedding social interaction knowledge into the generation of the wearer’s mesh. A primary advantage of DDPM is its capability to incorporate signals, represented as $\mathbf{c}$, that guide the trajectories of the conditional denoising process $\epsilon_\psi(\mathbf{z}_t, t, \mathbf{c})$. In our approach, we use a pre-trained VAE encoder specifically to encode the interactee’s pose $\mathbf{P^i}$ as a key conditioning element. During training, ground truth poses are utilized, while at test time, we employ the state-of-the-art EgoHMR algorithm \cite{Zhang2023ProbabilisticHM}. Additionally, we incorporate scene conditioning by compressing the point-cloud representation $\mathbf{S}$ of the environment with an MLP network $\tau_{\eta}$, as interactions are shaped by both social cues and the surroundings.

To inject these embedded conditions into the transformer-based denoiser, we concatenate different conditions and apply cross-attention~\cite{Chen2022ExecutingYC} with the denoiser latent. 
Then, the conditional objective is defined as follows:
\begin{equation}
    \mathcal{L} = \mathbb{E}_{\epsilon \sim \mathcal{N}(0,\mathbf{I}),t, \bold{P_i}, \bold{S}}[ \| \epsilon - \epsilon_{\psi} (\mathbf{z}_t,t,\mathcal{E}_\phi(\bold{P_i}),\tau_\eta(\bold{S}) \|^2_2 ]\,,
    \label{eq:diffusion_opt}
\end{equation}
where $\mathcal{E}_\phi$ and $\tau_\eta$ stay frozen.

\section{Experiments}
\label{ref:experiments}
In this section, we validate our model against the state of the art and showcase the qualitative results of our approach. Additionally, we perform several ablation studies highlighting the impact of social interactions. Below, we define the dataset and the reference metrics.

\paragraph{Dataset.} We employ the EgoBody~\cite{Zhang2021EgoBodyHB} dataset to assess our technique, the sole available egocentric and social dataset featuring an environment and body meshes defined by parametric body models~\cite{Loper2015SMPLAS} for both the wearer and the interactee. 
The dataset is recorded using Microsoft Hololens2~\cite{hololens} and contains 125 sequences at 30 fps. 
The total number of frames is 220k, but, as in~\cite{Zhang2021EgoBodyHB}, we selectively use frames where the interactee is within the wearer's field of view. 
The training set comprises 90000 images, the validation set consists of 23000 images, and the test set includes 62000 images. 
These recordings occur in 15 indoor scenes, each accompanied by 3D representations, which are recorded using an iPhone 12 Pro Max running a 3D Scanner App (for additional information, refer to \cite{Li2022EgoBodyPE}).

To enhance comparability, we also include the GIMO dataset~\cite{Zheng2022GIMOGH} and benchmark SEE-Me without Int.ee—meaning the variant of SEE-ME based solely on the scene, as GIMO contains only single-person videos. Although this specific setup excludes social cues, we conduct this evaluation to further assess our overall framework.
This dataset features egocentric views in single-body environments, and we adapt our model accordingly. It has been acquired from 11 different subjects performing actions in 19 3D scenes, scanned using an Apple iPhone 13 Pro Max. 
The dataset includes up to 125,400 egocentric images captured at 30 fps with a Hololens2 and 217 trajectories captured by an Inertial Measurement Units (IMU) system. In total, GIMO contains approximately 129,000 frames

\begin{table*}[t]
\centering
\caption{Egocentric Pose Estimation Results. We activate and deactivate Scene and interactee conditioning to assess their contribution. In each version, our framework improves SoA performances by a large margin. We obtain the best results when the scene and interactee conditioning are present. 
% \FG{I feel it'd be more straightforward if we wrote "SEE-ME w/o Int.ee" and "SEE-ME w/o Scene", if it fits}
}
\resizebox{0.8\textwidth}{!}{
\begin{tabular}{c|cc|cccc}
\cline{1-7} \textbf{Models}
 & \multicolumn{2}{c|}{\textbf{Conditioning}} & \multicolumn{4}{c}{\textbf{EgoBody \cite{Zhang2021EgoBodyHB} Dataset}} \\
 & Scene & Interactee & MPJPE $(mm)$ & \begin{tabular}[c]{@{}c@{}}Orientation\\ Error  \end{tabular} & \begin{tabular}[c]{@{}c@{}}Translation\\ Error $(mm)$ \end{tabular} & \begin{tabular}[c]{@{}c@{}}Acceleration\\ Error  $(mm/s^2)$ \end{tabular} \\ \hline
\multicolumn{1}{l|}{EgoEgo~\cite{Li2022EgoBodyPE}} & & & 268 & \textbf{0.40} & 207 & 10.8 \\ 
\hline
% \multicolumn{1}{l|}{SEE-ME Inter.} & & \ding{51} & 138 & 0.51 & 174 & 3.07 \\
% \multicolumn{1}{l|}{SEE-ME Scene} & \ding{51} & &  130 & 0.51 & 171 & 2.69 \\ 
\multicolumn{1}{l|}{SEE-ME w/o Scene} & & \ding{51} & 138 & 0.51 & 174 & 3.07 \\
\multicolumn{1}{l|}{SEE-ME w/o Int.ee} & \ding{51} & &  130 & 0.51 & 171 & 2.69 \\ 
\multicolumn{1}{l|}{SEE-ME} & \ding{51} & \ding{51} &  \textbf{126} & 0.48 & \textbf{164} & \textbf{2.67} \\ 
\hline
\end{tabular}}
\label{tab:egopose}
\end{table*}

\paragraph{Baselines.} To estimate the wearer's pose from a first-person perspective, we evaluate our model against EgoEgo~\cite{Li2022EgoBodyPE}, the existing state-of-the-art model designed for estimating the wearer's pose from an egocentric video.
EgoEgo is a probabilistic model based on DDPM~\cite{Ho2020DenoisingDP} that leverages head poses as conditioning to generate a plausible pose. 
The model consists of two modules. Firstly, it estimates the head pose by using two sub-modules. 
The first module estimates the global orientation and translation of the camera based on optical flow. The second module involves head pose estimation using DROID-SLAM~\cite{teed2021droid}. 
By combining these estimates, the output is the global position of the head, comprised of pose, orientation, and trajectory. 
During the second phase, the estimated head is employed as a conditioning factor for a DDPM model, facilitating the generation of realistic poses. 
We employ EgoEgo's~\cite{Li2022EgoBodyPE} configuration, where body poses are represented relative to the initial pose.
Concerning the scene alignment, we adopt an approach in line with EgoHMR\cite{Zhang2023ProbabilisticHM}.

\paragraph{Metrics.} We consider the evaluation metrics commonly employed in the current literature on egocentric human pose estimation~\cite{Zhang2023ProbabilisticHM, Li2022EgoBodyPE, Zhang2021EgoBodyHB}.\\
Specifically, we evaluate the accuracy of our model by computing the error on the keypoints extracted from the SMPL body representation, ignoring the mesh form factor. 
We use SMPL due to its widespread adoption for representing virtual humans. 
The added realism brought by meshes allows for augmenting the fidelity of interactions, for example, by modeling collisions. 

The Mean Per Joint Position Error (MPJPE) measures the average Euclidean distance between the predicted and ground truth 3D positions of individual joints across a sequence of frames:
\begin{equation}
\text{MPJPE} = \frac{1}{J \times T} \sum_{j=1}^{J}  \sum_{t=1}^{T} \| \mathbf{p}_{j,t} - \mathbf{g}_{j,t} \|^2.
\end{equation}

We measure the Orientation Error using the Frobenius norm of the 3x3 rotation matrix of the reference joint $\mathbf{A}_{pred}$ predicted and ground truth $\mathbf{A}_{GT}$, expressed as follows:
\begin{equation}
    \text{Orientation Error} = || \mathbf{A}_{pred} - \mathbf{A}_{GT}^{-1} - \mathbf{I} ||_2.
\end{equation}

To evaluate the error of the generated motion translation, we use the Euclidean distance between the predicted trajectory $\mathbf{r}^{pred}$ and the ground truth  $\mathbf{r}^{GT}$: 
\begin{equation}
    \text{Translation Error} = \frac{1}{T} \sum_{t=1}^{T} || \mathbf{r}^{pred}_t - \mathbf{r}^{GT}_t||^2.
\end{equation}

We then compute the acceleration for predicted poses $\mathbf{a}^{pred}$ and ground truth $\mathbf{a}^{GT}$, and we define the Acceleration Error as their Euclidean distance.
We employ millimeters ($mm$) for MPJPE and Translation, millimeters per second squared ($mm/s^2$) for Acceleration, and the Frobenius norm between the rotation matrices for Orientation.

\subsection{Comparison with SOTA}

We quantitatively assess the camera wearer's pose using the results from the EgoBody~\cite{Zhang2021EgoBodyHB} dataset. As the GIMO~\cite{Zheng2022GIMOGH} dataset lacks any interactee, we benchmark only the SEE-ME variant, focusing solely on the scene. Furthermore, we conduct several studies to explore the impact of social relation proxies and their effects.

\begin{figure*}[h!t]
    \centering
    \resizebox{\textwidth}{!}{
    \begin{tabular}{ccc}
            \multicolumn{2}{c}{\includegraphics[width=1\textwidth]{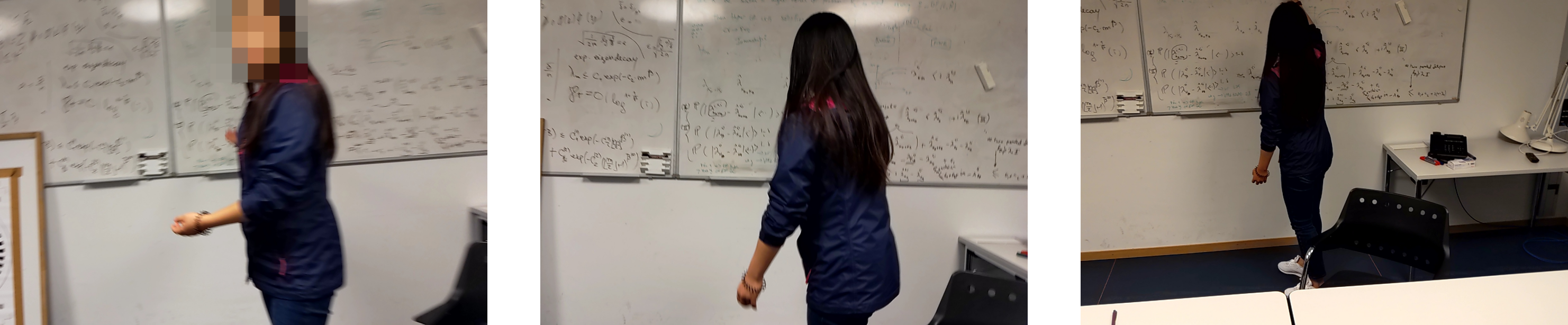}}
         &  \includegraphics[width=.5\textwidth]{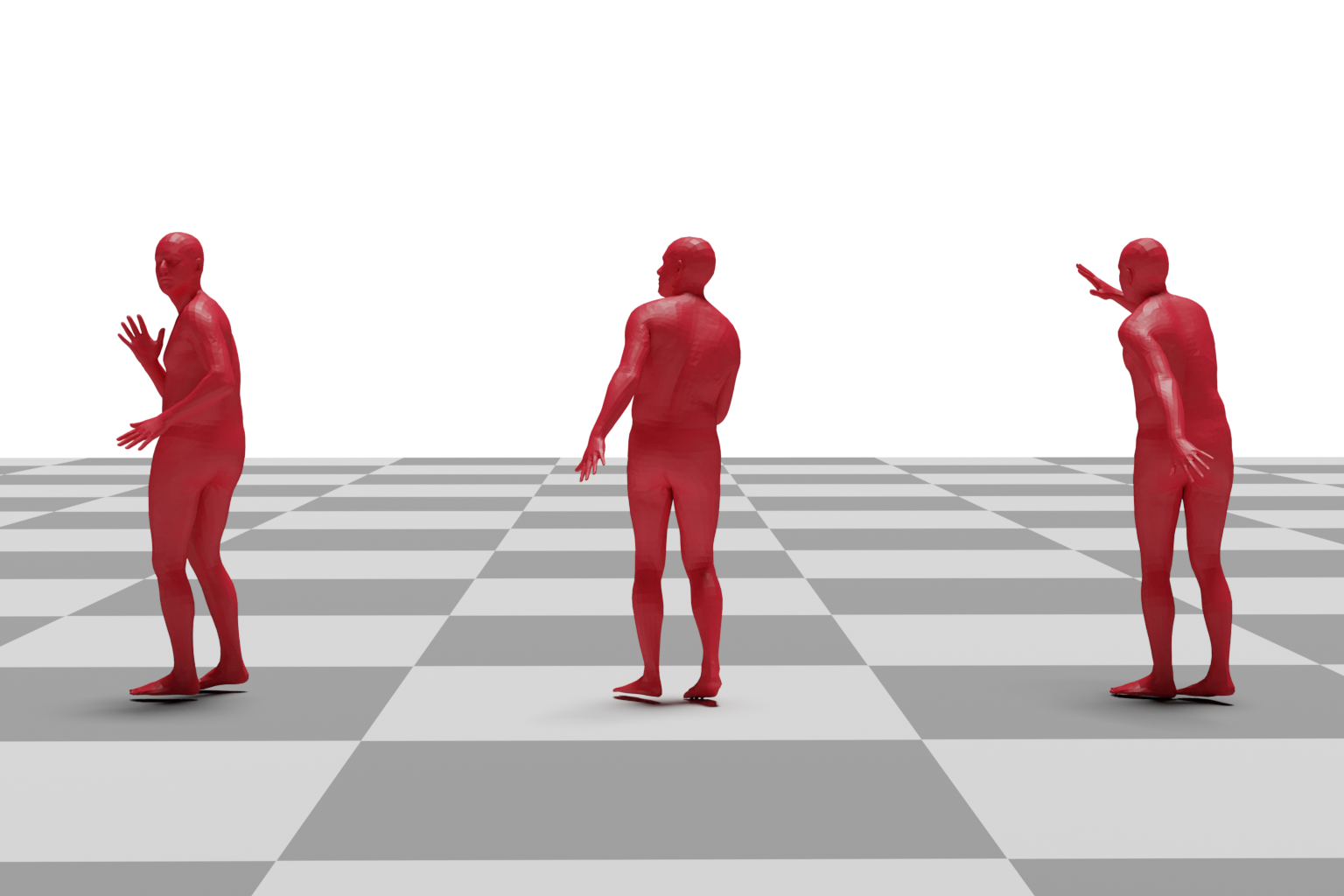} \\
          \multicolumn{2}{c}{(a) Video stream of the wearer's egocentric view, not used in the model.} &
          (b) Wearer's view of the interactee, extracted from the video stream.
         \\
        \includegraphics[width=.5\textwidth]{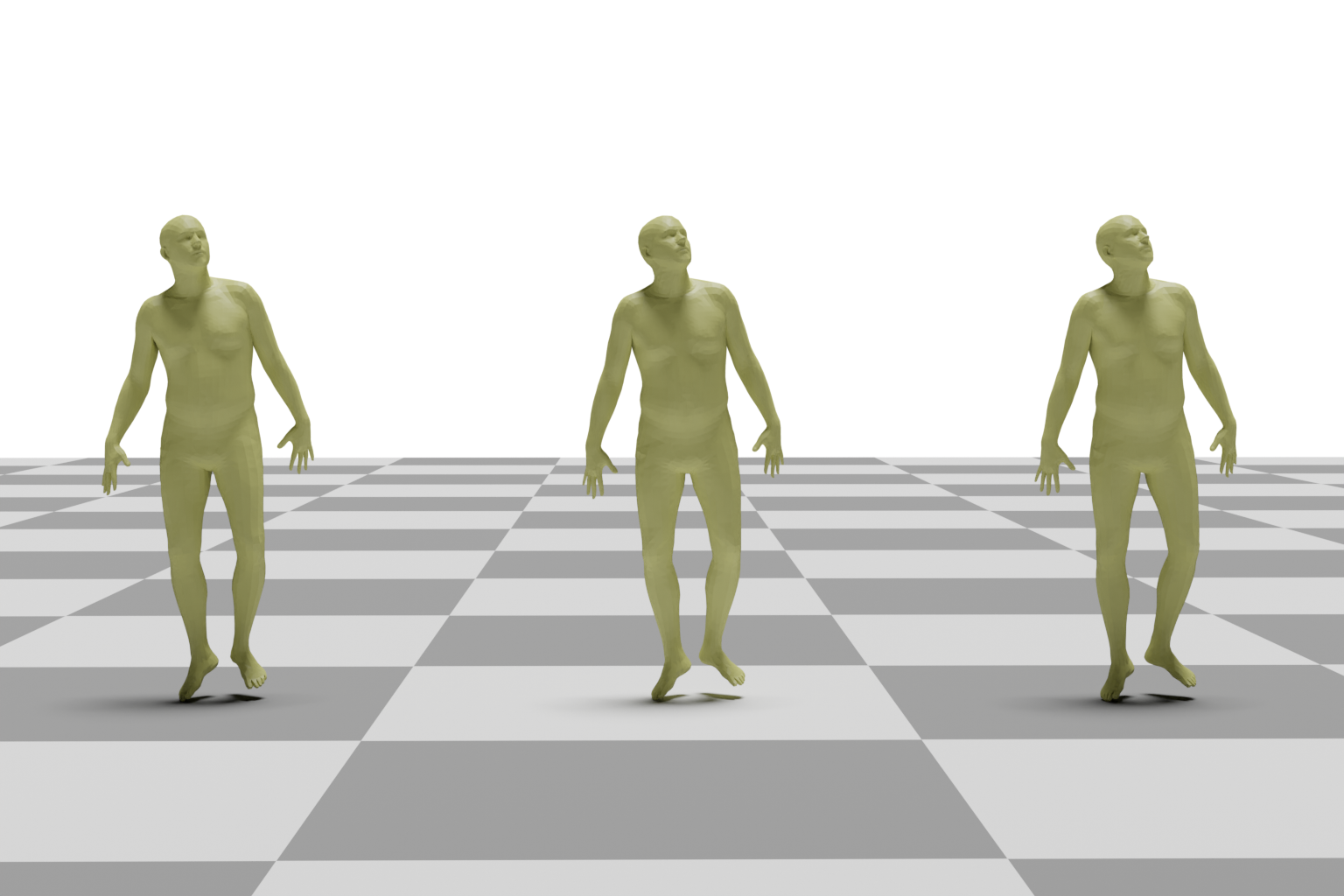}
         &  \includegraphics[width=.5\textwidth]{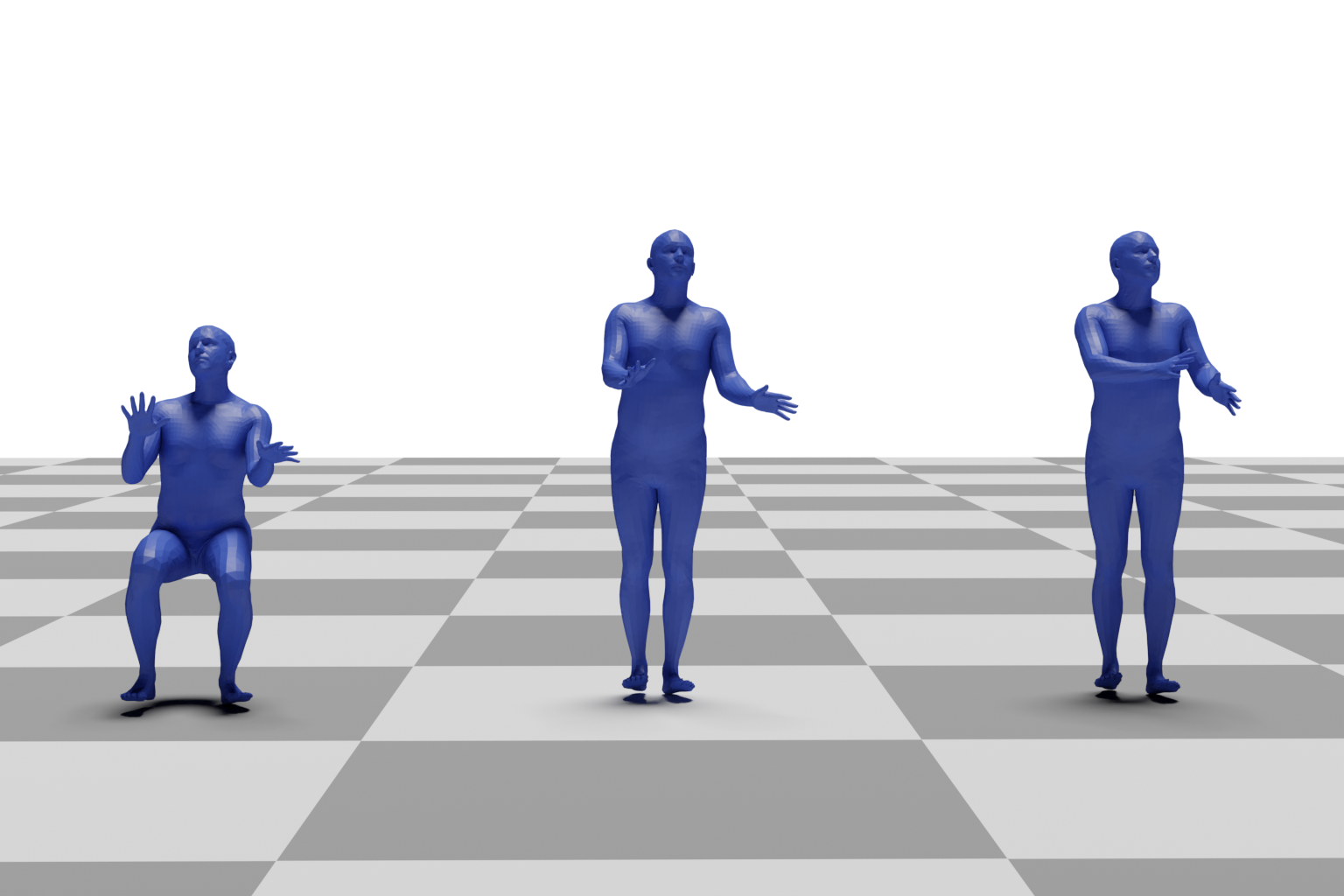} & \includegraphics[width=.5\textwidth]{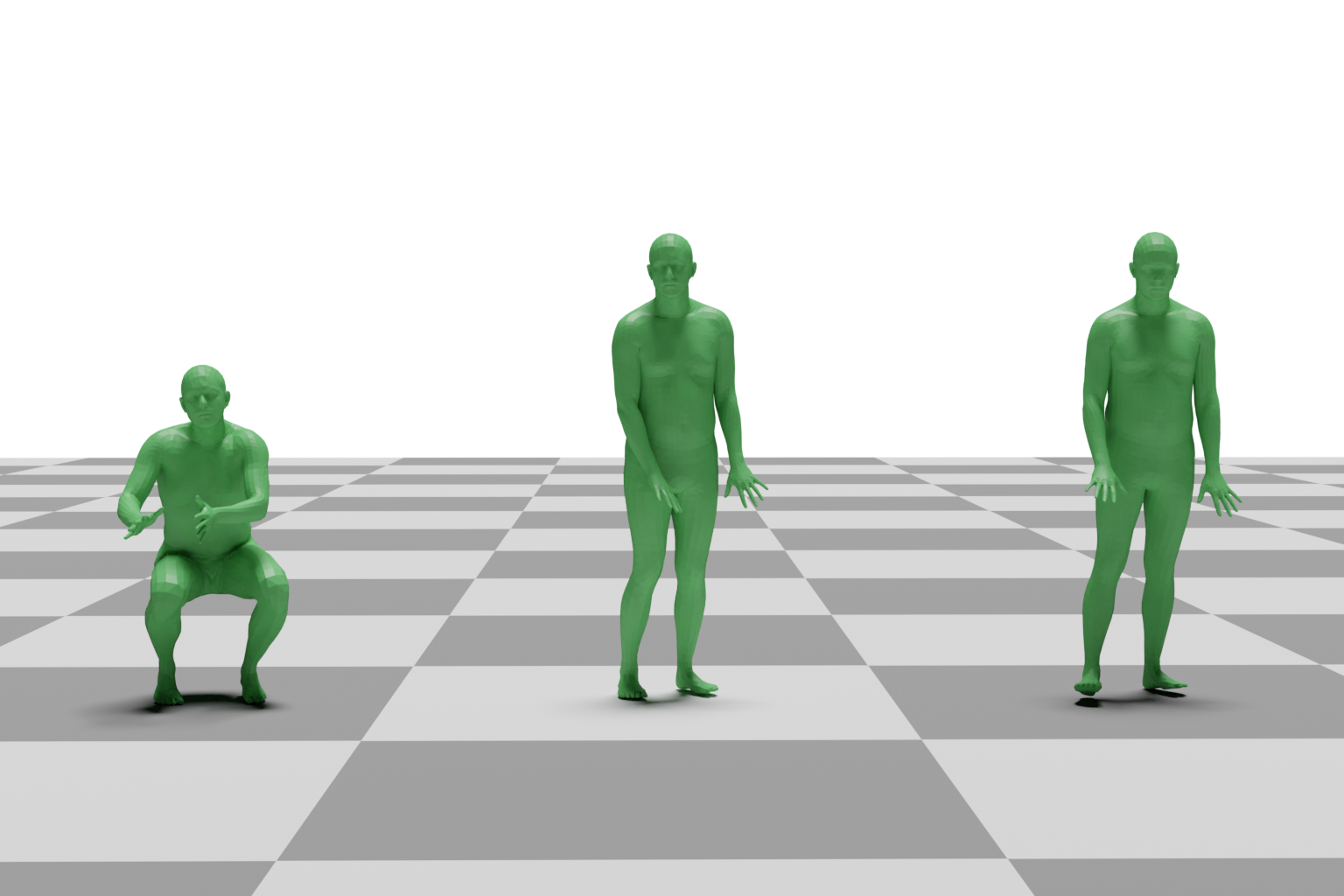} \\
          (c)  Prediction of the wearer's pose by EgoEgo. &
          (d) Prediction of the wearer's pose by SEE-ME. &
          (e)  Ground truth wearer pose.
         
    \end{tabular}}

    \caption{Front view of 3 frames extracted from an egocentric sequence.
    We compare SEE-ME (blue) with EgoEgo (yellow), and the ground truth (green).
    In red we have the interactee's poses, extracted from the egocentric video sequence next to it, but not used in our model.
    }
    \label{fig:results}
    \vspace{-.5cm}
\end{figure*}

\paragraph{Social egocentric human pose estimation.} 
Table~\ref{tab:egopose} compares our model quantitatively against the current leading technique~\cite{Li2022EgoBodyPE}. We retrained the HeadNet module of their model on the EgoBody~\cite{Zhang2021EgoBodyHB} dataset, specifically targeting head rotation and translation distance estimation. The rest of the model's modules, initially trained on a large-scale motion capture dataset like AMASS \cite{Mahmood2019AMASSAO}, were frozen during this process.

We improve the performance of MPJPE from 268mm to 126mm over the state-of-the-art~\cite{Li2022EgoBodyPE} yielding a 53\% enhancement in MPJPE, 21\%  in Translation, and a significant 75\% in Acceleration. 
The impact of the interactee is evident, even when the wearer is not engaged in active interaction with it.
Furthermore, compared to \cite{Li2022EgoBodyPE}, conditioning solely on the interactee enhances the performance from 268mm to 138mm in MPJPE, giving a 49\% increase and a substantial 72\% improvement in the Acceleration error between predicted and ground truth joints. 

Predicting the wearer's pose conditioning on the scene alone allows a direct comparison with EgoEgo, and our model proves again to boost performances.
Going from 268mm to 130mm, we get a 51\% MPJPE improvement, predicting the wearer's pose from the scene alone
We get 17\% and 75\% improvements in translation and acceleration errors, respectively.

\paragraph{Egocentric human pose estimation.}
\begin{table*}[!ht]
\centering
\resizebox{0.65\textwidth}{!}{
\begin{tabular}{l|ccc}
\hline
Models     & MPJPE $(mm)$ & Orientation Error & Translation Error $(mm)$\\ \hline
PoseReg~\cite{yuan2019ego}   & 189   & 1.51              & 1528              \\
Kinpoly-OF~\cite{luo2021dynamics} & 404   & 1.52              & 1739              \\
EgoEgo~\cite{Li2022EgoBodyPE}     & 152   & 0.67              & \textbf{356}               \\ \hline
% SEE-ME Scene-Only   & \textbf{141}   & \textbf{0.61}              & 843               \\ \hline
SEE-ME w/o Int.ee   & \textbf{141}   & \textbf{0.61}              & 843               \\ \hline
\end{tabular}}
\caption{Quantitative results on GIMO~\cite{Zheng2022GIMOGH} dataset. 
}
\label{tab:gimo}
\vspace{-.3cm}
\end{table*}

To reinforce the effectiveness of our approach, we further assess SEE-ME using the GIMO dataset, as there is a scarcity of 3D social egocentric environmental datasets aside from EgoBody.
Thus we set to assess the quality of SEE-ME by the sole consideration of the scene.
%, neglecting the absent social cues.
% Thus we readjust our task from human pose estimation from social cues to only consider the room we are in.
The results shown in Table \ref{tab:gimo} are comparable to EgoEgo and they should be paralleled with SEE-ME w/o Int.ee in the prior experiment, as the interactee is absent in GIMO.

SEE-ME w/o Int.ee outperforms EgoEgo on the orientation estimation but it yields a larger translation error as EgoEgo leverages information from SLAM.

Overall, on the general MPJPE performance, SEE-ME w/o Int.ee outperforms EgoEgo by 7.2\%, which reasserts the quality of the proposed model, across different settings.

\subsection{Qualitative results}

We show the results obtained from SEE-ME qualitatively, comparing them to the state-of-the-art~\cite{Li2022EgoBodyPE}. In Figure~\ref{fig:results}, the input footage (a) is processed to extract the interactee's pose by EgoHMR~\cite{Zhang2023ProbabilisticHM} (b), which is then ingested by SEE-ME, alongside the encoding of the scene point cloud. The qualitative reconstructions match the quantitative evaluation of Table~\ref{tab:egopose} as SEE-ME (d) adheres better to the ground truth, than EgoEgo (c) does. E.g.\ the actor stands up in (d) to attend to the interactee.

\begin{figure}
    \centering
    \includegraphics[width=0.42\textwidth, bb=0 0 1500 1050]{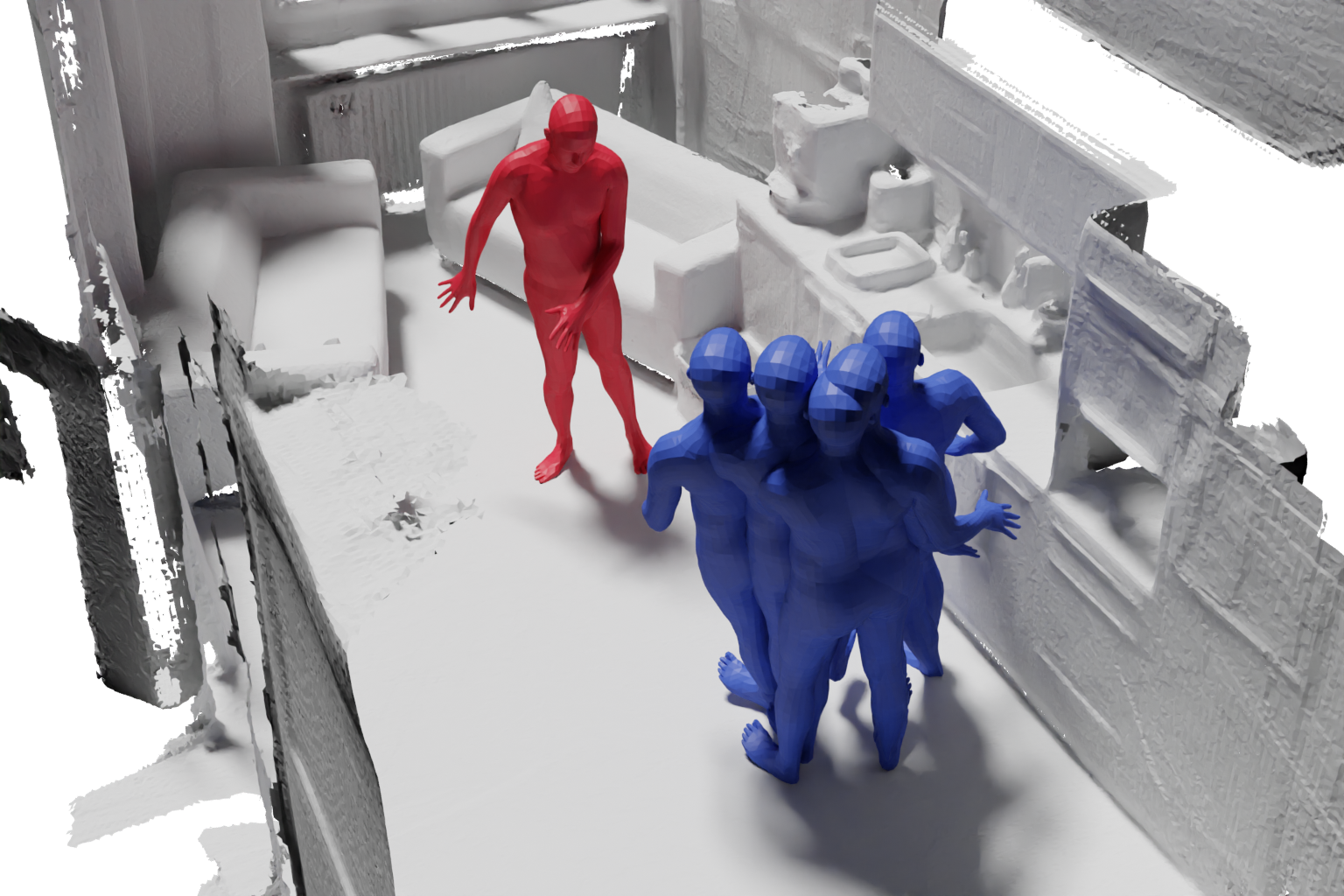}
    \caption{The interactee (red) influences the wearer's motion (blue).}
    \label{fig:interactee-vs-wearer}
\end{figure}

Figure \ref{fig:interactee-vs-wearer} depicts a different qualitative study case: the interactee's pose (red) conditions the generation of the ego actor (blue), which SEE-ME successfully positions in front of it, taking plausible poses while conversing.

\vspace{-1mm}
\subsection{Ablation studies}

\paragraph{Social Interaction ablation.}
We explore the influence of social relations through ablation studies to gain a deeper understanding of the interactee's impact. 
To quantify interaction, we initially employ a straightforward proxy: the distance between individuals. 
This is achieved by categorizing distances into three ranges based on the root joint. 
As illustrated in Table \ref{tab:distance}, the proximity of the interactee and wearer correlates with more pronounced effects on the final results. 
Notably, the most substantial improvements are observed in MPJPE (by 6\%), Translation Error (by 9\%), and Acceleration (by 23\%) when considering proximate interactions. 

The diminished performance in Translation and Acceleration beyond the 2-meter range is attributed to the wearer moving toward the interactee. 
As individuals draw closer, there is a heightened relative acceleration and translation, making it more susceptible to errors.

\begin{table*}[!ht]
\centering

\resizebox{.65\textwidth}{!}{
\begin{tabular}{l|cccc}
\cline{1-5}
\multicolumn{1}{c|}{\begin{tabular}[c]{@{}c@{}}Wearer-Interactee\\ Distance Threshold ($m$) \end{tabular}} & MPJPE ($mm$) & \begin{tabular}[c]{@{}c@{}}Orientation\\ Error\end{tabular} & \begin{tabular}[c]{@{}c@{}}Translation\\ Error ($mm$)\end{tabular} & \begin{tabular}[c]{@{}c@{}}Acceleration\\ Error ($mm/s^2$)\end{tabular} \\ 
\hline
\multicolumn{1}{c|}{-} & 126 & \textbf{0.48} & 164 & 2.67 \\
\hline
\multicolumn{1}{c|}{$d > 2$} & 132 & \textbf{0.48} & 183 & 2.76 \\ 
\multicolumn{1}{c|}{$1 < d < 2$} & 128 & 0.49 & 161 & 2.73 \\ 
\multicolumn{1}{c|}{$d < 1$} & \textbf{119} & \textbf{0.48} & \textbf{156} & \textbf{2.06} \\ 
\hline
\end{tabular}}
\caption{Ablation study on the interpersonal distance between wearer and interactee. Conditioning on the interactee's pose works best when in close proximity.}
\label{tab:distance}
\end{table*}

The investigation in Table \ref{tab:degree} aims to ascertain mutual gaze fixation between two individuals using information derived from their respective head rotation matrices.
Mutual gaze, a critical component of non-verbal communication, can indicate interpersonal engagement. 
The proposed method relies on converting rotation matrices to Euler angles, extracting gaze directions, and evaluating the angular deviation between these vectors.
We employ $30^{\circ}$ and $60^{\circ}$ thresholds to determine whether two individuals are making eye contact. 
We look at two data subsets for each threshold: one satisfying the condition and the other not.
Remarkably, both thresholds yield improved results when satisfied, affirming the initial intuition. 
The most substantial improvement occurs when individuals are directly looking at each other ($30^{\circ}$), leading to enhancements in MPJPE from 127mm to 117mm (by 8\%), from 175mm to 137mm in Translation Error (by 22\%), and also enhancing the Orientation Error (14\%), and Acceleration (by 9\%) for when the condition not being met.

\begin{table*}[!h]
\centering

\resizebox{0.65\textwidth}{!}{
\begin{tabular}{l|c|cccc}
\cline{1-6}
\multicolumn{1}{c|}{\begin{tabular}[c]{@{}c@{}}Wearer-Interactee\\ looking at each other\end{tabular}} & FOV & MPJPE & \begin{tabular}[c]{@{}c@{}}Orientation \\ Error\end{tabular} & \begin{tabular}[c]{@{}c@{}}Translation\\ Error\end{tabular} & \begin{tabular}[c]{@{}c@{}}Acceleration\\ Error\end{tabular} \\
\hline
\multicolumn{1}{c|}{-} & \multicolumn{1}{c|}{-} & 126 & 0.48 & 164 & 2.67 \\
\hline
\multicolumn{1}{c|}{No} & & 127 & 0.48 & 173 & 2.71 \\
\multicolumn{1}{c|}{Yes} & \multicolumn{1}{c|}{\multirow{-2}{*}{60}} & \textbf{123} & \textbf{0.48} & \textbf{157} & \textbf{2.57} \\ 
\hline
\multicolumn{1}{c|}{No} & & 127 & 0.50 & 175 & 2.79 \\
\multicolumn{1}{c|}{Yes} & \multicolumn{1}{c|}{\multirow{-2}{*}{30}} & \textbf{117}   & \textbf{0.43} & \textbf{137} & \textbf{2.54} \\ 
\hline
\end{tabular}}
\caption{Ablation study on gaze directions. By considering an angle of 60 and 30 degrees, we asses if the wearer and the interactee are looking at each other. If this is the case, the conditioning boots improve the performance even more.}
\label{tab:degree}
\end{table*}
Finally, we examine a scenario in which the wearer possesses knowledge of the interactee's future movements. 
This is achieved by conditioning on future frames rather than on the present. 
As indicated in Table \ref{tab:future}, the MPJPE and the Translation Error exhibit improvements of 2\% and 22\%, respectively. 
The latter's enhancement can be attributed to the wearer's anticipation of the interactee's movements, reducing Translation error.

\begin{table*}[!h]
\centering

\resizebox{0.65\textwidth}{!}{
\begin{tabular}{l|cccc}
\cline{1-5}
\multicolumn{1}{c|}{\begin{tabular}[c]{@{}c@{}}Wearer's input = present\\ Interactee's input = $x$\end{tabular}} & MPJPE & \begin{tabular}[c]{@{}c@{}}Orientation\\ Error\end{tabular} & \begin{tabular}[c]{@{}c@{}}Translation\\ Error\end{tabular} & \begin{tabular}[c]{@{}c@{}}Acceleration\\ Error\end{tabular} \\ \hline
\multicolumn{1}{c|}{$x$  = present} & 126   & \textbf{0.48} & 164 & \textbf{2.67} \\
\hline
\multicolumn{1}{c|}{$x$ = future} & \textbf{123} & \textbf{0.48} & \textbf{128} & 3.40 \\
 \hline
\end{tabular}}
\caption{Ablation study on wearer poses conditioned on the interactee's present and future ones. Even a little glance into the future reduces the MPJPE.}
\label{tab:future}
%\vspace{-.5cm}
\end{table*}

\vspace{-4mm}
\paragraph{Implementation details.}
Our framework consists of three main components: the encoder, the decoder, and the denoiser. 
The reconstruction phase, which has 24 million parameters, is handled by the encoder and decoder, while the generation phase involves the denoiser (9 million parameters) and the frozen decoder.
Each component has 9 layers 4 heads, and a latent space whose dimensionality is 256.
The encoder and the decoder both use standard transformer layers.
Based on \cite{Zhang2022MotionDiffuseTH}, the denoiser combines classical self-attention on the latent vector with linear cross-attention between the latent vector and textual input.
The batch size is 64 for the first phase and 128 for the second one, and the AdamW optimizer is used with a learning rate of $10^{-4}$.
The diffusion steps are set to 1000 and 20 during training and inference.
We train on 8 Tesla V100 GPUs for 3k epochs for both phases.

\vspace{-5mm}
\paragraph{Limitations and Future Work.}

We do not inject any social relations knowledge externally but leave the model free of learning from the dataset. 
This means that the quality of the predictions is highly dependent on the training data distribution.
While we recognize the presence of social relations in the utilized dataset, EgoBody\cite{Zhang2023ProbabilisticHM}, we acknowledge that these relationships could be more effectively leveraged in more diverse and socially engaging datasets, such as EgoHumans~\cite{Khirodkar2023EgoHumansAE} for which currently both mesh recovery and 3D point cloud are not available. 
Additionally, we recognize the potential to enhance our model by exploiting large-motion datasets such as AMASS, and possibly others such as KIT, during the VAE training for the reconstruction phase. This would allow us to capture movements beyond the constraints of a single dataset.
Currently, our model works without taking any wearer's input, but considering its egocentric observed body parts as additional data could be a research path to explore, and which could further improve our performances.

\vspace{-2mm}
\section{Conclusions}
\label{sec:conclusions}

In conclusion, accurately determining the 3D pose of the camera wearer in egocentric video sequences is pivotal for advancing human behavior modeling in virtual and augmented reality applications. 
Despite the challenges posed by limited visibility, SEE-ME has demonstrated that the pose of the wearer can be reconstructed to an improved level of accuracy.
By utilizing a latent probabilistic diffusion model, our approach integrates conditioning techniques that effectively capture both social interactions and the surrounding environment. Moreover, it is straightforward and does not require any extra preprocessing compared to other methods that utilize localization techniques. This development shows great potential for improving the realism and precision of egocentric video-based human behavior modeling, particularly for applications in augmented reality/virtual reality (AR/VR) and embodied AI.

\vspace{-3mm}
\paragraph{Acknowledgements}
This project was supported by PNRR MUR project PE0000013-FAIR and from the Sapienza grant RG123188B3EF6A80 (CENTS). We acknowledge the CINECA award under the ISCRA initiative for the availability of high-performance computing resources and support.

% \newpage

%%%%%%%%% REFERENCES
{\small
\bibliographystyle{ieee_fullname}
\bibliography{main}
}

\end{document}